\pdfoutput=1

\documentclass[11pt]{article}

\usepackage[final]{acl}

\usepackage{times}
\usepackage{latexsym}

\usepackage[T1]{fontenc}

\usepackage[utf8]{inputenc}

\usepackage{microtype}

\usepackage{inconsolata}

\usepackage{graphicx}

\usepackage{booktabs}
\usepackage{amsmath}
\usepackage{natbib}
\usepackage{listings}
\usepackage{xcolor} 
\usepackage{datetime}

\newcommand{\claude}{Claude-3.5S}
\newcommand{\gptt}{GPT-3.5T}
\newcommand{\gpto}{GPT-4o}

\newcommand{\gemini}{Gemini-1.5P}
\newcommand{\llama}{Llama-3 70B}
\newcommand{\sllama}{Llama-3 7B}

\title{Towards Reproducible LLM Evaluation: Quantifying Uncertainty in LLM Benchmark Scores}

\author{
 \textbf{Robert E. Blackwell\textsuperscript{1}},
 \textbf{Jon Barry\textsuperscript{2}},
 \textbf{Anthony G. Cohn\textsuperscript{3,1}}
\\
\\
 \textsuperscript{1}The Alan Turing Institute, \\
 \textsuperscript{2}The Centre for Environment Fisheries and Aquaculture Science,\\ 
 \textsuperscript{3}School of Computer Science, University of Leeds.
\\
 \small{
   \textbf{Correspondence:} \href{mailto:rblackwell@turing.ac.uk}{rblackwell@turing.ac.uk}
 }
}

\begin{document}
\maketitle
\begin{abstract}
Large language models (LLMs) are stochastic, and not all models give deterministic answers, even when setting $temperature$ to zero with a fixed random $seed$.
However, few benchmark studies attempt to quantify uncertainty, partly due to the time and cost of repeated experiments.
We use benchmarks designed for testing LLMs' capacity to reason about cardinal directions to explore the impact of experimental repeats on mean score and prediction interval.
We suggest a simple method for cost-effectively quantifying the uncertainty of a benchmark score and make recommendations concerning reproducible LLM evaluation.
\end{abstract}

\section{Introduction}

As \emph{Generative Artificial Intelligence (GenAI)} systems become prevalent, it is natural to want to
assess their capabilities and compare their performance.
\emph{Large Language Models (LLMs)} \citep{devlin-etal-2019-bert,brown2020language}, such as Claude Sonnet and GPT-4o
are examples of so-called \emph{Foundation Models} \citep{bommasani2021opportunities} that
generate textual responses to prompts,
having been trained on very large
corpora.
Vendors compete to provide ever more capable LLMs, making claims and counter-claims about
model performance across a variety of tasks including general knowledge and reasoning (e.g. see recent product launches from XAI\footnote{\url{https://x.ai/blog/grok-2}, accessed October 2024.} and OpenAI\footnote{\url{https://openai.com/index/gpt-4o-mini-advancing-cost-efficient-intelligence/}, accessed October 2024.}).

LLM evaluation is a burgeoning field \citep[for a survey see][]{chang2024survey}.
LLM benchmarks consisting of question and answer pairs, are widely used to assess performance and provide leader boards comparing state-of-the-art, frontier
models (e.g. LMSYS Chatbot Arena\footnote{\url{https://huggingface.co/spaces/lmsys/chatbot-arena-leaderboard}, accessed October 2024.}).
Specialist benchmarks are emerging to assess model performance across
diverse domains including coding \citep[e.g.][]{zhuo2024bigcodebench}, medical \cite[e.g.][]{cai2024medbench} and legal \citep[e.g.][]{guha2024legalbench}.

However, LLMs are stochastic systems \citep{bender2021dangers} that may generate non
deterministic answers.
Despite calls for reproducibility in LLM evaluation \citep[e.g.][]{burnell2023rethink}, few studies
try to quantify uncertainty.
Attempts have been made to reduce answer
variability (for example, by fixing the random number $seed$, prompt
engineering \citep{sahoo2024systematic} or changing the sampling strategy \citep[e.g., nucleus sampling, ][]{Holtzman2020The}, but at the time of writing, none of these approaches is reliable for all APIs and models.

We note that reproducibility and determinism are not always desirable
properties of an AI system, especially if novel or artistic solutions
are sought. The subject of creativity in AI has a long history, for
example see \citet{boden2004creative}. However, in this paper we are
primarily interested in benchmarks with questions that can either
be answered correctly or incorrectly and can thus test
specific capabilities of an LLM.

Reasons for non-determinism in LLMs include the use of probabilistic random sampling, the unpredictable order of subsystem execution in parallel systems and differences in floating point arithmetic implementation \cite{hagmann2023towards}. LLMs generate text based on probabilities derived from their training data. At each step in the generating process, the model predicts the next word (or token) by sampling a probability distribution of next token likelihood. Whilst random number generators can be seeded, coordination of these seeds is difficult in a distributed system where the order of execution of parallel elements is unpredictable. This is especially the case in shared, cloud infrastructure which may consist of heterogeneous hardware.

Commercial LLM \emph{Application Programming Interfaces (APIs)} can be costly and are often rate-limited, restricting the throughput and number of experiments
that can reasonably be undertaken as part of a benchmark study.
In this paper, we explore how the size of benchmarks, the model parameters, and the number of experimental repeats affect uncertainty. We conclude by making practical recommendations for cost-effective LLM sampling, evaluation, and benchmarking.

\section{Related Work}

Lin et al (\citeyear{lin2024generatingconfidenceuncertaintyquantification}) differentiate uncertainty from confidence: the former refers to the “dispersion” of the potential predictions for a fixed input, and the latter refers to the confidence of a particular prediction.

Lower temperature settings are associated with more deterministic responses, whereas higher settings promote creativity. In their experiments, Renze and Guven
(\citeyear{renze2024effectsamplingtemperatureproblem}) showed that changes in temperature from 0.0 to 1.0 did not have a statistically significant effect on the problem solving performance of the LLMs tested.
Renze and Guven
(\citeyear{renze2024effectsamplingtemperatureproblem}) also showed that uncertainty reduces with temperature and
recommend setting an LLM’s sampling temperature to 0.0 to make results more deterministic. Patel et al. (\citeyear{Patel2024.07.22.24310824}) report similar results for clinical tasks.




\section{Methods}

To explore the stochastic nature of LLMs, we present questions to
OpenAI GPT-3.5 Turbo 0125 (\gptt),
OpenAI GPT-4o  2024-05-13 (\gpto),
Meta Llama 3 70B Instruct (\llama),
Meta Llama 3 7B (\sllama),
Google Gemini 1.5 Pro  preview-0409 (\gemini),
and
Anthropic Claude 3.5 Sonnet 20240620 (\claude). The OpenAI models are hosted on Microsoft Azure (with content filtering switched off), \llama\ on Microsoft Azure AI, \gemini\ on Google Vertex
and \sllama\ on Ollama on a Mac M2.
Each question is posed as a prompt in a separate chat completion.

We study two representative, qualitative spatial reasoning benchmarks developed by Cohn and Blackwell (\citeyear{cohn2024evaluatingabilitylargelanguage}).
The first benchmark (\emph{Small}) consists of 100 simple questions with cardinal directions as answers, e.g. \emph{``You are watching the sun set. Which direction are you facing?''}. The second benchmark (\emph{Large}) consists of 5760 templated questions with cardinal or intercardinal directions as answers, e.g. \emph{``You are walking south along the east shore of a lake; in which direction is the lake?''}. We run the experiment as described by Cohn and Blackwell (\citeyear{cohn2024evaluatingabilitylargelanguage}) except that we first run with default model settings (not explicitly setting $seed$, $temperature$ or $top\_p$).
We then run with $temperature$ = 0 and a fixed $seed$ = 123.We use the system prompt \emph{``You are a helpful assistant. I will give you a question about directions. The answer is either north, south, east, or west. Please only reply with the answer. No yapping.''} for \emph{Small} and \emph{``You are a helpful assistant. I will give you a question about directions. The answer is either north, south, east, west. north-east, north-west, south-east or south-west. Please only reply with the answer. No yapping.''} for \emph{Large}. We vary the number of repeats from 1 to 30 to explore the impact of repetition on mean benchmark score and compute prediction interval as a measure of uncertainty:

Let \( q \) be the number of benchmark questions, \( n \) the number of times the experiment is repeated and \( X_{i,j} \in \{0, 1\}\) the score for the \( i \)-th question in the \( j \)-th repeat (0 incorrect, 1 correct).
The mean score for a single repeat \( j \), with \( q \) questions is given by (Eq. \ref{eq:e1}), and the mean score over \( n \) repeats, by (Eq. \ref{eq:e2}).

\begin{equation}
\bar{x}_j = \frac{1}{q} \sum_{i=1}^{q} X_{i,j}
\label{eq:e1}
\end{equation}

\begin{equation}
\bar{x} = \frac{1}{n} \sum_{j=1}^{n} \bar{x}_j
\label{eq:e2}
\end{equation}

The prediction interval for a future observation of the mean \( \bar{x}' \) over \( n' \) repeats is (\ref{eq:e3}).

\begin{equation}
 \bar{x} \pm t_{\alpha/2, n-1} \cdot s \cdot \sqrt{\frac{1}{n}+\frac{1}{n'}}
\label{eq:e3}
\end{equation}

where
\( s \) is the standard deviation of \( \bar{x} \) and
\( t_{\alpha/2, n-1} \) is the critical value from the Student's \( t \)-distribution for a confidence level \( 1 - \alpha \) and \( n-1 \) degrees of freedom (we use \( \alpha\) = 0.05 for a 95\% probability).

\citet{tian2022methods} give a short history of prediction intervals dating back to \citet{fisher1935fiducial}. Generally, they are used for a single future observation ($n' = 1$), but, here, we use them for a future mean.
Note that a prediction interval is wider than its corresponding confidence interval. A prediction interval estimates the range where future observations or means of observations will likely fall, while a confidence interval estimates the range where the true population parameter (e.g. mean) lies based on sample data \citep{chiolero2012meta}. We are interested in reproducibility of benchmark scores and so we use prediction intervals where $n' = n$.

Finally, to explore whether there are differences in the same model provided by different vendors, we run the \emph{Small} cardinal reasoning benchmark ($temperature$ = 0, $seed$ = 123, $n$ = 90) using the Microsoft provided Azure OpenAI API and then the OpenAI provided API. We perform a two-sample t-test to determine whether the results are statistically significantly different at the $\alpha$ = 0.05 significance level.

\section{Results}

\begin{figure*}[htb!]
\centering
\includegraphics[width=0.99\textwidth]{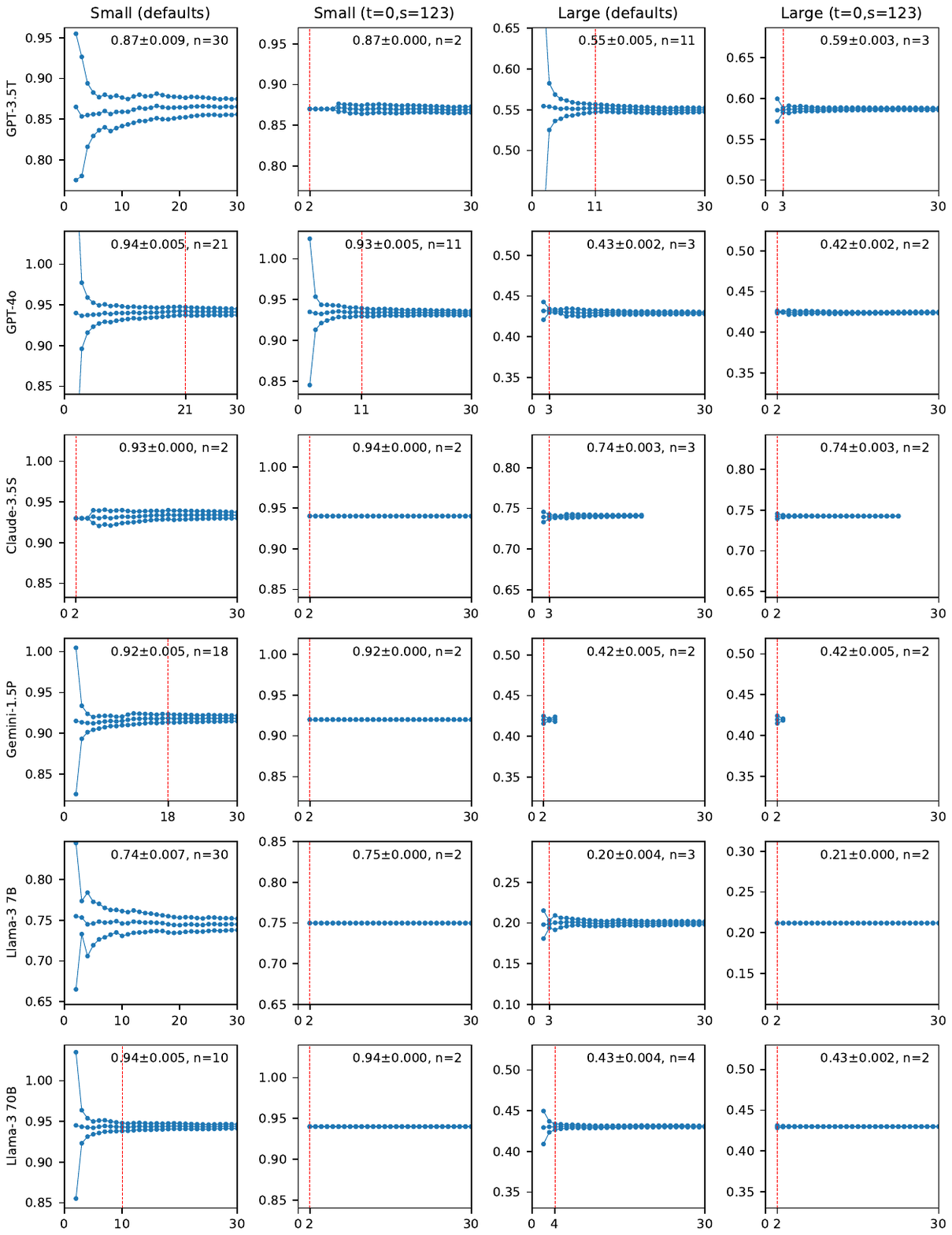} 
\caption{Prediction interval by repeat for each model tested with \emph{Small} and \emph{Large} benchmarks. The red dashed line indicates the repeat
at which the prediction interval width falls below 0.01. For \claude\ and \gemini\ applied to \emph{Large} the total number of repeats tested is fewer than 30 owing to practical constraints.}
\label{pi}
\end{figure*}

For all models tested, when using the default model settings, the
\emph{Large} benchmark gave a prediction interval width $<$ 0.01 after fewer
repeats than the \emph{Small} benchmark (Figure \ref{pi}).  All tests
except \gptt\ with \emph{Large} gave a prediction interval width $<$ 0.01 in
four or fewer repeats.

For both \emph{Large} and \emph{Small} benchmarks, setting temperature
to 0.0 with a fixed $seed$ gave a prediction interval width $<$ 0.01 after
less than or the same number repeats than when using default settings.
In all cases except two (\gptt\ and \sllama\ with \emph{Small}
(defaults)), the prediction interval width reduces to $<$ 0.01 within $n$ = 30
repeats.

When comparing results for \gptt\ from the Microsoft Azure OpenAI API and the OpenAI API applied to the \emph{Small} dataset, the mean scores were 0.833 and 0.840 respectively ($n$ = 90).
A two-sample t-test suggests that the results are statistically significantly different ($t$ = 2.51, $p$ = 0.013).

\section{Discussion}

Although we found no combination of $temperature$ and $seed$ or
$top\_p$ and $seed$ that gave deterministic answers for all models and
benchmarks, setting $temperature$ to 0 and fixing $seed$ (where available)
reduced variability for most models and experiments, as expected. Unless creativity
of answers is a goal, or there is a requirement to test the default
settings of models, it seems sensible to set $temperature$ to 0 with a
fixed $seed$ and thus reduce the number of experimental repeats
required to achieve a desired prediction interval width.
Evaluators might chose to tune parameters such as $temperature$ or
$top\_p$ to maximise performance on a model by model or task by task basis, but this
could be time consuming and we prefer to compare
models using consistent settings.

\sllama\ always gave deterministic results when setting the $seed$ to a fixed value.
Hagmann et al.(\citeyear{hagmann2023towards})
note the challenges of reproducibility when using distributed, parallel architectures and out-of-order execution. \sllama\ is the only model tested that was run locally rather than in the cloud
and we suggest that this lack of distributed parallelism allows the random number generator to be consistently seeded, resulting in predictable execution patterns and deterministic results.

It is important to document the precise API, model, version, and parameters being used to conduct a benchmark.
Our results show that \gptt\  via the Microsoft Azure OpenAI API
and \gptt\ via the OpenAI API, gave statistically
significantly different results for
the
\emph{Small} benchmark. Although model versions appear to be consistent in both the Azure OpenAI API and \mbox{OpenAI} API documentation,
the two APIs are different and use different message formats.
At the time of writing there were multiple versions of \gptt\ available
(including gpt-3.5-turbo-0125,
gpt-3.5-turbo-1106,
gpt-3.5-turbo-instruct-0914,
gpt-35-turbo-16k-0613, and
gpt-35-turbo-0613) and some APIs (e.g. Microsoft Azure OpenAI API) can be set to auto-update, causing possible confusion. Our results
also show that model parameters such as $temperature$ and $seed$ can also change the mean score of a model, e.g. setting $temperature$ to 0 and $seed$ to 123 for
\gptt\ applied to \emph{Large} increased its score from 0.55 to 0.59 (Figure \ref{pi}).

Inference APIs can be expensive and rate limited, making experimental repeats costly and time-consuming.
Our results suggest that when setting $temperature$ to 0.0 with
a fixed $seed$, it is rarely necessary to conduct more than three repeats to achieve a prediction interval width of $\leq$ 0.01.

\section{Conclusions}

LLMs are stochastic and not all LLMs provide deterministic answers. 
However,
setting $temperature$ to 0.0 with a fixed $seed$ appears to reduce answer variability for many models and reduces the possibility of outlier benchmark scores.

Evaluators should take steps to quantify uncertainty in LLM benchmark scores.
One way to quantify uncertainty is to provide results in
the form $\bar{x} \pm \epsilon$, using prediction intervals (Eq.~\ref{eq:e3}); the number of
experimental repeats can be increased incrementally until the prediction interval width is below a desirable threshold (perhaps 0.01). Larger benchmarks with more questions reduce the variability of the mean score.

The experimental conditions including benchmark size, model, version, API, parameters, number of experimental repeats, and date of the experiment should be carefully documented, as should the
method used to quantify uncertainty.

\section{Limitations}

This study tested six representative LLMs from four vendors, but there are now many LLMs that could be tested with varying architectures, weights, context windows and parameters.

This study tested two qualitative spatial reasoning benchmarks of different size, question style and complexity, but there are now many benchmarks that could be tested. Different benchmarks are designed to test different knowledge and reasoning tasks. It is possible that some benchmarks may
require more experimental repeats for the prediction interval width to reduce acceptably.

We have not looked at the impact of prompt engineering on answer variability.

We have not looked at the impact of linguistic variation (e.g. tense, person form, vocabulary or language) on answer variability.

This study used temperature based sampling and does not consider nucleus ($top\_p$) sampling. Preliminary work by the authors suggest that nucleus sampling can also be used to reduce the variability of benchmark score and that the methods described herein for quantifying uncertainty are equally applicable.

LLMs are black box systems and we have no way of knowing precisely how
models will perform for future requests. The underlying hardware and software systems may be subject to change without notice. LLM technology is evolving
rapidly, and new sampling methods may change the stochasticity of LLMs.

In this paper we have used frequentist statistics to model uncertainty, but it would also be interesting to explore Bayesian statistical approaches.

\section{Acknowledgements}

This work was supported by the Fundamental Research priority area of
The Alan Turing Institute.

We also thank Microsoft Research -- Accelerating Foundation Models
Research Program for the provision of Azure resources to test
Microsoft-provided models.

\bibliography{references}

\appendix
\onecolumn

\section{Supplementary Information}
\label{sec:appendix}

\begin{figure*}[htb!]
\centering
\includegraphics[width=0.99\textwidth]{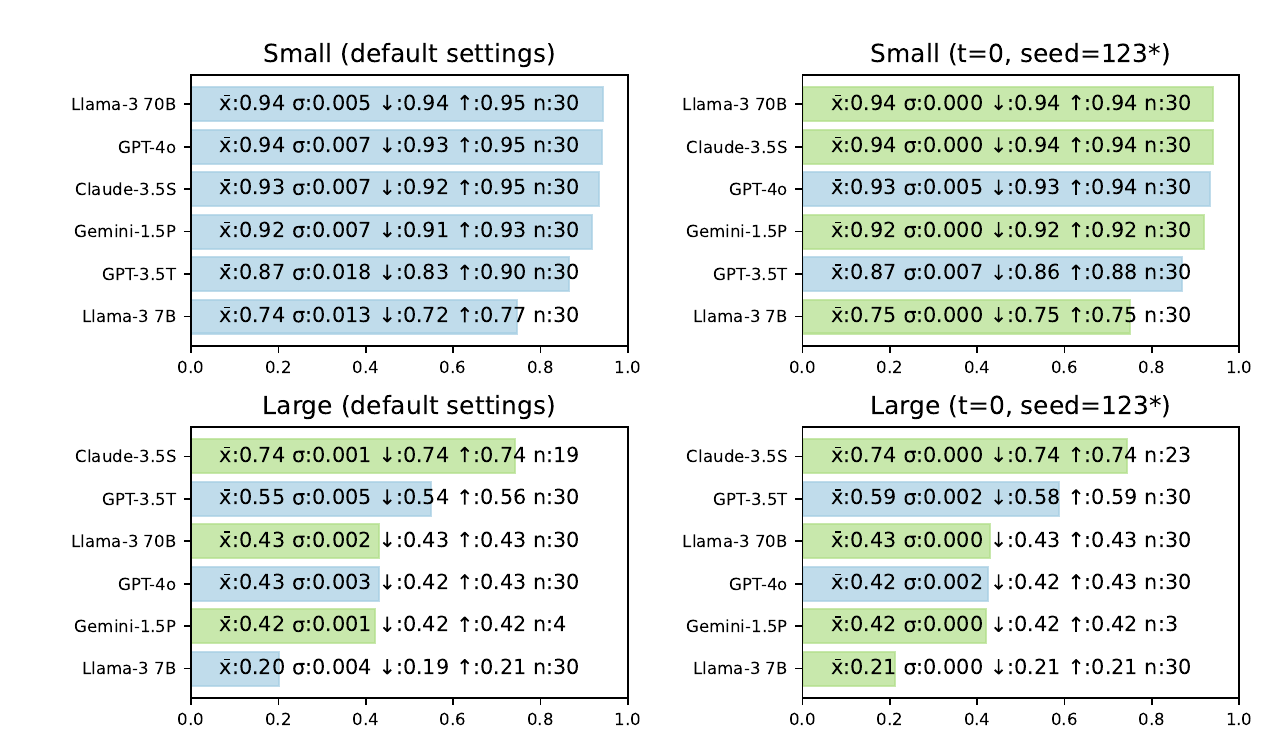} 
\caption{Mean score by model for
the
\emph{Small} benchmark (top) and
the
\emph{Large} benchmark (bottom). Models use default settings on the left and $temperature$ = 0 with a fixed $seed$ on the right. The text on the bars shows mean score ($\bar{x}$), standard deviation ($\sigma$), worst score ($\downarrow$), best score ($\uparrow$), and the number of experimental repeats ($n$). Green bars indicate range $<$ 0.01. *Note that the $seed$ parameter cannot be set for \claude.}
\label{cosit}
\end{figure*}

Figure \ref{cosit} shows summary information for each of the models tested.
\\

The following Python code was used to compute the 95\% prediction interval.

\begin{lstlisting}
import numpy as np
import scipy.stats as stats

def prediction_interval(samples, confidence=0.95):

    mean = np.mean(samples)
    std_dev = np.std(samples, ddof=1)  # sample standard deviation

    n = len(samples)

    t_crit = stats.t.ppf((1 + confidence) / 2, df=n - 1)

    margin_of_error = t_crit * std_dev * np.sqrt(2 / n)

    lower_bound = mean - margin_of_error
    upper_bound = mean + margin_of_error

    return lower_bound, upper_bound

\end{lstlisting}

\begin{figure*}[htb!]
\centering
\includegraphics[width=0.99\textwidth]{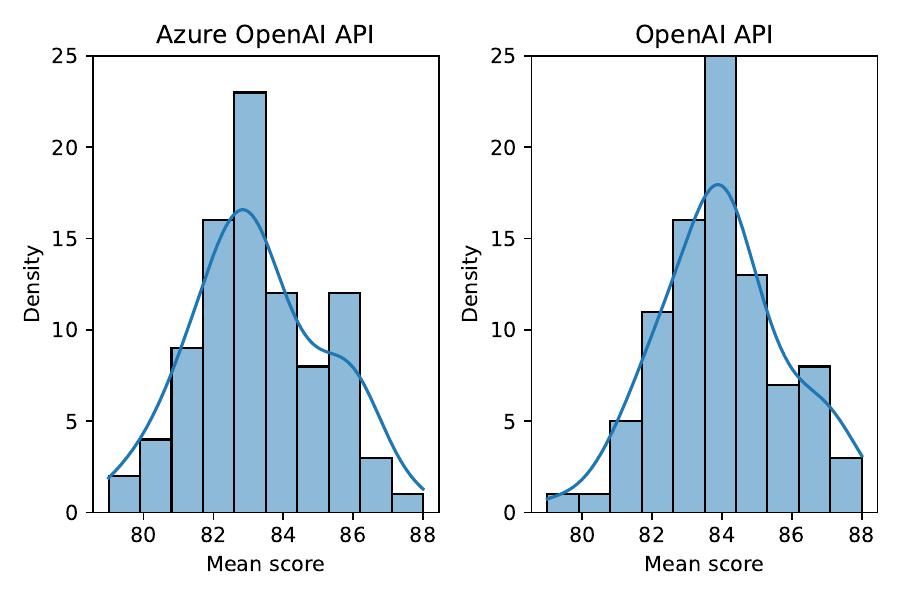} 
\caption{Histogram of mean score for
the
\emph{Small} benchmark for \gptt\ using the Azure OpenAI API (left) and the OpenAI API (right). The two distributions are statistically significantly different ($t$ = 2.51, $p$ = 0.013, $n$ = 90), indicating possible differences in the hosting of the model. }
\label{signif}
\end{figure*}

\begin{figure*}[htb!]
\centering
\includegraphics[width=0.99\textwidth]{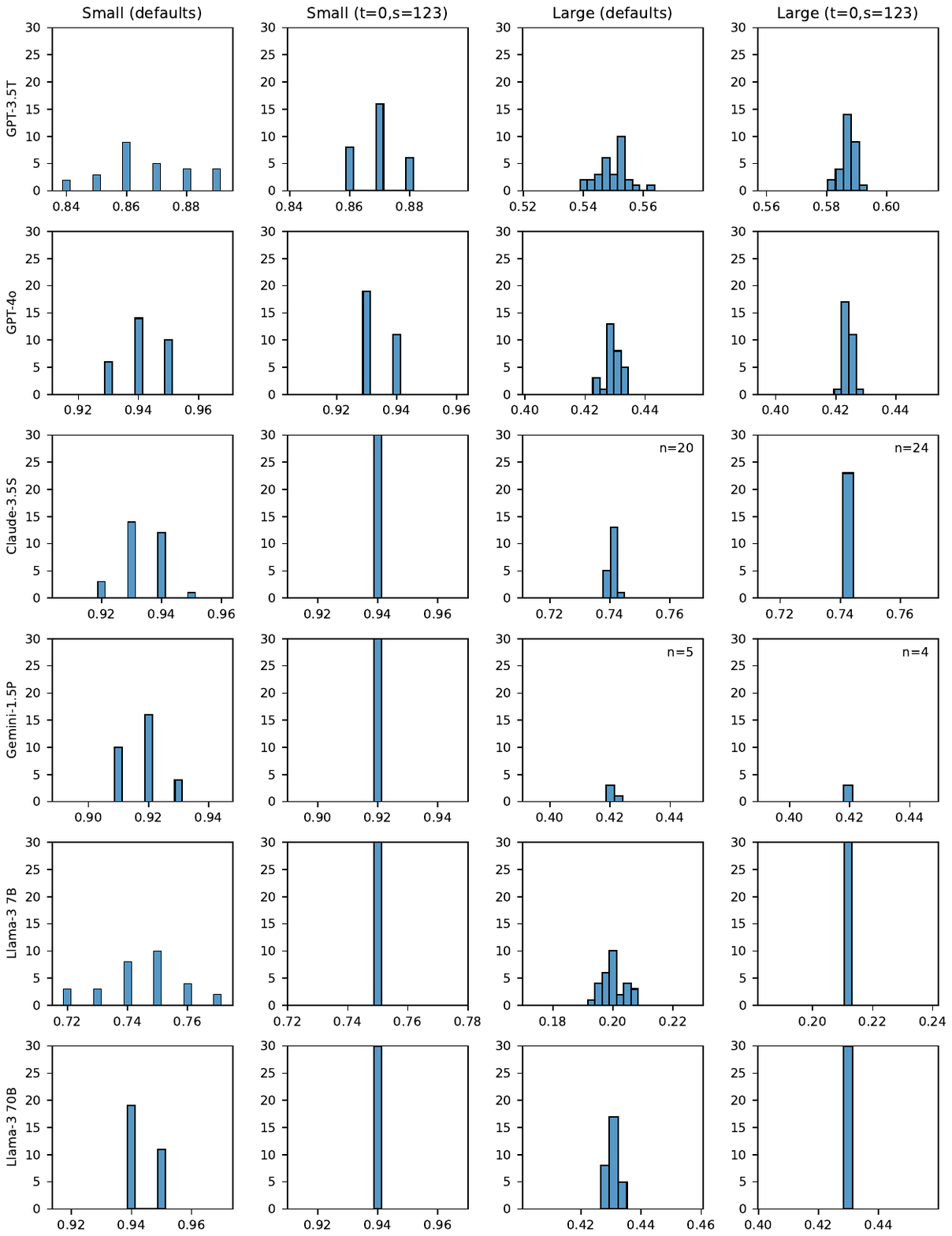} 
\caption{Histogram of mean score for for each model tested with \emph{Small} and \emph{Large} benchmarks. In most cases the number of repeats, n=30, but in some cases n is fewer owing to practical constraints;
in these cases n is specified in the top right of the plot. }
\label{histograms}
\end{figure*}


\end{document}